\documentclass[10pt,twocolumn,letterpaper]{article}

\usepackage{iccv}
\usepackage{times}
\usepackage{epsfig}
\usepackage{graphicx}
\usepackage{amsmath}
\usepackage{amssymb}

\usepackage[breaklinks=true,bookmarks=false]{hyperref}

\iccvfinalcopy 

\def\iccvPaperID{****} 

\ificcvfinal\fi

\begin{document}

\title{Spatio-Temporal Video Representation Learning for \\AI Based Video Playback Style Prediction}

\author{
Rishubh Parihar \thanks{Equal Contribution}\\
Indian Institute of Science,\\
Bangalore, India.\\
{\tt\small parihar.rishubh@gmail.com}

\and
Gaurav Ramola \footnotemark[1]\\
Samsung India Research Institute,\\
Bangalore, India.\\
{\tt\small g.ramola@samsung.com}
\and
Ranajit Saha\\
Microsoft Corporation,\\
Hyderabad, India.\\
{\tt\small rs.ranajitsaha@gmail.com}
\and
Ravi Kini\\
Samsung India Research Institute,\\
Bangalore, India.\\
{\tt\small abcdravi@gmail.com}
\and
Aniket Rege\\
Univ. of Washinton,\\
Seattle, USA.\\
{\tt\small aniketrg7@gmail.com}
\and
Sudha Velusamy\\
Samsung India Research Institute,\\
Bangalore, India.\\
{\tt\small sudha.v@samsung.com}
}

\maketitle

\begin{abstract}
Ever-increasing smartphone-generated video content demands intelligent techniques to edit and enhance videos on power-constrained devices. Most of the best performing algorithms for video understanding tasks like action recognition, localization, etc., rely heavily on rich spatio-temporal representations to make accurate predictions. For effective learning of the spatio-temporal representation, it is crucial to understand the underlying object motion patterns present in the video. In this paper, we propose a novel approach for understanding object motions via motion type classification. The proposed motion type classifier predicts a motion type for the video based on the trajectories of the objects present. 
Our classifier assigns a motion type for the given video from the following five primitive motion classes: linear, projectile, oscillatory, local and random. 
We demonstrate that the representations learned from the motion type classification generalizes well for the challenging downstream task of video retrieval. Further, we proposed a recommendation system for video playback style based on the motion type classifier predictions.

\end{abstract}

\section{Introduction}


An increasing volume of smart-phones with high-quality cameras in recent years has led to a meteoric rise in the amount of video content captured and shared on social media platforms such as Tiktok, YouTube, Facebook, Instagram, SnapChat, ShareChat etc. This trend has fostered the need for automated video analysis tools that can aid the user to edit  videos with ease on mobile devices, on-the-fly.

\begin{figure}[t]
   \centering
   \includegraphics[width=7.8cm, height=4cm]{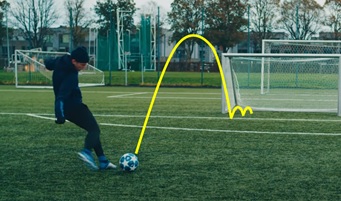}
    \caption{Visualizing an example motion trajectory of a ball }
    \vspace{-0.1cm}
    \label{Fig:Football}
\end{figure}

Videos contain rich information embedded in both spatial and temporal dimensions, which together capture the overall dynamics of the scene. Learning meaningful spatio-temporal representation is at the core of most video analysis tasks like video retrieval, action recognition, temporal and spatial action localization, object motion analysis, video captioning, and modelling of human-object interactions. 
There is a fundamental need for methods to learn generalized spatio-temporal representations that can work effectively for multiple downstream tasks. 
One of the popular approaches is to train a model for video action recognition and obtain the implicitly learned video representation \cite{carreira2017quo} \cite{tran2018closer}. Recently, many self-supervised methods have been proposed, where a deep network is trained for an auxiliary pre-text task to learn rich spatio-temporal representations. 

Object motion understanding is crucial to learn rich spatio-temporal representations as it provides insights about the natural motion pattern of objects in the world and how they interact with other objects in the scene~\cite{zhang2020learning}. For instance, consider the example of a video where a person is shooting a ball towards the goalpost as shown in Fig.~\ref{Fig:Football}. Analysing the motion of the ball during this action will provide insight about the most likely motion of the soccer ball: just after kicking, the ball will follow a projectile motion in the air, and after dropping on floor the ball will bounce a few times. This motion pattern of a relatively common occurrences in everyday life is extremely complex to model in a mathematical or mechanical sense as it comprises, for instance in the above example, movement of the player's body and real world forces (friction, air drag) at play. 

In this work, we present a method of analysing the underlying directional information of object motions that occur in real-world human actions like kicking, walking, jumping, clapping, etc., by estimating the object motion type in a video. As it is difficult to jointly model motions of all the objects in the scene, we focus only on the dominant motion in the video. To this end, we have formulated a classification problem, to classify the directional motion pattern into one of the defined classes. Based on our internal study on action classes present in popular video dataset HMDB$51$ \cite{kuehne2011hmdb}, we have defined five primitive motion type classes: \emph{linear, projectile, oscillatory, local and random.} According to us, most of the real world human actions can be assigned to one of the above defined motion types. For instance: \emph{walking, running, bike-riding} have a linear motion type as the dominant motion, \emph{kicking, cartwheel} makes projectile motion, and \emph{talking, chewing, smoking} have a local motion type. All the motion patterns having periodic motion are considered under oscillatory class, for example, \emph{pushup and exercise.} The actions which do not lie into any of these categories were assigned the class random. To our knowledge, there is no open-source video dataset currently available with motion type labels for videos. To this end, we have added motion type annotations to the HMD51~\cite{kuehne2011hmdb} dataset for training the motion classifier. The motivation of this work is to address the following:\emph{1) Is it possible for a neural network model to perform well on the task of motion type classification? 2) What internal feature representations does the model learn in this process? 3) Are these learned features generalize well on other downstream video understanding tasks?} We have tried to answer these questions throughout this paper by training a CNN model for motion type classification and analyzing its learned features through general video analysis tasks like video retrieval.

We also demonstrate an exciting use-case of the above-presented motion type classification method: video playback style recommendation, which boosts the overall aesthetics of the videos. A few common playback styles include: Reverse (temporally reversing the video), Loop (repeating the video in a loop), Boomerang (playing a concatenated video of normal and reverse). Finding a suitable playback style is often a time-consuming process where a user manually applies each available playback style. This created a space to engineer automated tools for this problem. Our proposed solution tries to automate this process of playback style selection. More details for the design of this recommendation algorithm are presented in Sec. 3.2. 

Lastly, we show that through the proposed motion type classification, we are able to learn rich spatio-temporal representations that generalize well for other video analysis tasks such as video retrieval. In a subjective evaluation of the learned representations for video retrieval, we achieved promising results on the HMDB$51$ dataset. Furthermore, we made specific design choices to make the network efficient for mobile deployment. Our model for motion classification has inference time of $200ms$ for a $10$ second video clip on a Samsung S20 phone.

We summarize our major contributions as follows:
\begin{enumerate}
    \item A neural network for understanding object motion in videos by classifying object motion type into one of the five primitive motion classes: \emph{linear, projectile, oscillatory, local and random.}
    \item A light-weight network for video representation learning that is suitable for real-time execution on mobile devices.
    \item A recommender system to predict suitable video playback style for videos by analysing predicted object motion patterns
\end{enumerate}

\section{Related Works}
\begin{figure*}[t]
   \centering
   \includegraphics[width=16cm, height=5cm]{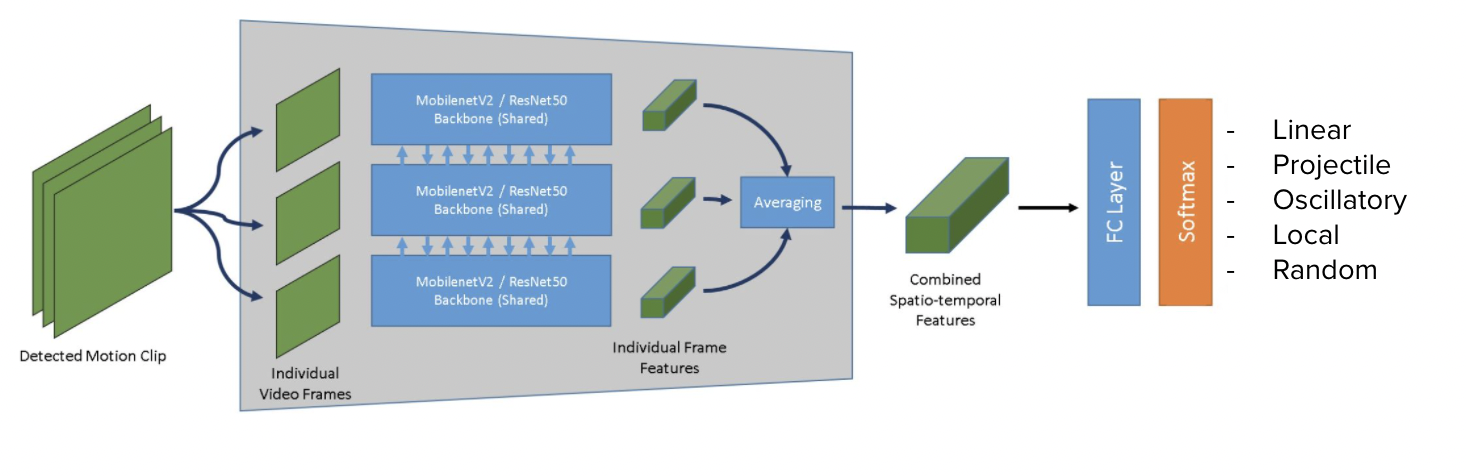}
    \caption{Overall network architecture for motion type classification. Given an input video, we divide it temporally into three segments and extract one central frame from each segment. These three frames are fed to the Feature Extractor Network, and the extracted features are then averaged to obtain a $1280$-dimensional (1280D) feature vector, which is used for motion type classification}
    \label{Fig:Architecture}
\end{figure*}
Video action recognition has been studied extensively by computer vision community. The success of video action recognition majorly depends on crafting the spatio-temporal features in the video representations. Traditionally, video features are extracted from optical-flow based motion information in the videos, e.g. Motion Boundary Histograms (MBH) \cite{dalal2006human} and trajectories \cite{wang2013action}, Histograms Of Flow (HOF) \cite{laptev2008learning} or spatiotemporal oriented filtering such as HOG3D \cite{klaser2008spatio}, Cuboids \cite{dollar2005behavior} and Spatiotemporal Oriented Energies (SOEs) \cite{feichtenhofer2015dynamically, derpanis2012action}.
The resounding success of Convolutional Neural Networks (CNNs) for image processing applications has caused its extension to video processing problems as well. Just like the spatial features, deep CNNs are also capable to extract accurate temporal information as well e.g. FlowNets \cite{dosovitskiy2015flownet,ilg2017flownet}. Both the temporal and spatial information are important in various video recognition tasks. Simonyan and Zisserman \cite{simonyan2014two} has proposed a two-stream CNN architecture to incorporate both spatial and temporal features of the videos. The spatial features are captured by passing the RGB frames of the videos and the temporal features are captured by extracting the flow frames. Several other works \cite{feichtenhofer2017spatiotemporal,feichtenhofer2016convolutional} have explored the different effective fusion options of two streams - flow and RGB streams.
The major bottleneck in two-stream networks as well as optical flow based methods is the optical flow extraction step as it consumes a lot of time and hence the inference time increases.

DMC-Net \cite{shou2019dmc} approximates the flow using a reconstruction loss and an adversarial loss jointly for the task of action classification. This model is two folds faster than the state-of-the-art methods and achieves accuracy close to the methods using optical flow information. The study of Tran \etal \cite{tran2015learning} shows the effectiveness of using $3D$-CNNs instead of $2D$-CNNs to model both spatial and temporal features together in a single branch. Although $3D$-CNNs produce promising results, it is much more expensive than $2D$-CNNs. Experiments by Xie \etal \cite{xie2018rethinking} showed that we can trade-off accuracy and speed by replacing some $3D$ conv layers by $2D$ convolutions. Having $3D$ conv layers at the higher layers and $2D$ conv layers at the lower part of the network is  faster and this configuration surprisingly has higher accuracy. They also propose separable $3D$-CNN (S3D) configuration which separates spatial and temporal $3D$ convolutions. MARS \cite{crasto2019mars} introduces the learning approaches to train $3D$-CNN operating on RGB frames which mimics the motion stream. It eradicates the need of flow extraction during the inference time. Frame sampling from videos is also an important part in video processing. Temporal Segment Network (TSN) \cite{wang2016temporal} works on sparse temporal snippets. The videos are split into k chunks and a small snippet in chosen from each of the chunk. The chunks are processed individually and at the end the decisions are aggregated as per the consensus function to come to the final conclusion. TSN gives promising result for action recognition task. Lin \etal \cite{lin2019tsm} proposes a generic module called Temporal Shift module (TSM). It is a "plug and play" module in a network designed for video understanding task. TSM has high efficiency and high performance. It maintains the complexity of $2D$-CNN and performance of the $3D$-CNN. TSM facilitates the information exchange by shifting a part of the channels along temporal dimension.

Object motion pattern understanding is crucial for learning strong spatio-temporal features for downstream video analysis tasks ~\cite{zhang2020learning}. There are approaches which try to capture the object motions in the videos via learning flow features from the videos \cite{fan2018end,ng2018actionflownet}. These methods predict pixel-level feature maps for every time frame in the video, which essentially captures only local motion patterns.

Most of the methods discussed above are based on the supervised learning technique. But due to the scarcity of publicly available labeled dataset, it is difficult to train deep networks with supervised learning. Several Self-supervised methods \cite{benaim2020speednet,han2019video} for video tasks have been studied by the computer vision community. Qian proposed \cite{qian2020spatiotemporal} self-supervised Contrastive Video Representation Learning (\textit{CVRL}) method which uses the contrastive loss to map the video clips in the embedding space. It is desired that in the embedding space the distance between two clips from the same video is lesser than the clips from different videos. Jenni \etal \cite{jenni2020video} introduced a novel self-supervised framework to learn video representations which are sensitive to the changes in the motion dynamics. They have observed that the motion of objects is essential for action recognition tasks. In the proposed work, we build on the above intuition to show that a deep network can learn rich representations by training for motion classification. 

\section{Methodology}
Humans largely use primary motion cues like underlying object motion patterns to understand video semantics like actions or events in a scene. To perform well on video analysis tasks like action recognition and localization, the motion pattern representations require a semantic understanding of both the appearance and dynamics features of the video. We aim to learn rich spatio-temporal video representations through classification of the motion type based on the directional motion information present in the video. To this end, we trained a motion type classification model that classifies a video into one of the following five primitive classes we define: \emph{linear, oscillatory, local, projectile, and random}. We observed that the trajectories of most natural object motions that we encounter in the real-world can be categorized into the first four motion classes. As  it  is  difficult  to  jointly  model  motions  of all  the  objects  in  the  scene,  we  focus  only  on  the  dominant motion in the video. For instance, actions such as \emph{walk} and \emph{run} usually follow a linear trajectory and have a dominant linear motion. Many activities that we perform indoors have motion in only small local regions like \emph{eat, drink, chew, talk}. Some of the examples of actions having dominant oscillatory motion type are \emph{dribble, cartwheel and sit-up}. \emph{Catch, throw, golf} are examples for dominant projectile motion type. Actions which do not follow any of these directional patterns, are considered random, for instance \emph{dance and fight}. Some of the common real-world actions and their corresponding motion types are shown in Table~\ref{tab:playback_mapping}. To validate the quality of our learned representations, we used these representations for video retrieval task as explained in Sec.~\ref{subsec:vid_retrieval}. As there is no publicly available video dataset with motion type labels, we have annotated the HMDB$51$ dataset with motion type labels to obtain mHMDB51 dataset as seen in Sec.~\ref{subsec:dataset}. The core of our method is a Deep Convolutional Neural Network (Fig.~\ref{Fig:Architecture}), which is trained in a supervised fashion on mHMDB$51$ dataset for a five class motion-type classification problem.


\begin{table}[t]
\begin{center}
\caption{Mapping of real world actions to motion type and Video Playback Style in the mHMDB$51$ dataset.} \label{tab:playback_mapping}
\begin{tabular}{|l|l|l|}
  \hline
  Example Action   & Motion Type & Playback Style\\
  \hline
    Walk, Run & Linear	&	Reverse	\\
    Dive, Throw & Projectile	&	Boomerang	\\
    Eat, Clap & Local	&	Loop	\\
    PullUp & Oscillatory	&	Loop	\\
    Dance, Fight & Random	&	Random	\\
  \hline
\end{tabular}

\vspace{-0.4cm}
\end{center}
\end{table}

\subsection{Network Architecture}\label{sec:arch}
Most state-of-the-art networks for video representation learning and action recognition methods~\cite{hara2018can}~\cite{carreira2017quo} rely on $3D$ convolutions due to their ability to jointly learn both spatial and temporal features. However, $3D$ convolutions have significantly higher computational cost than $2D$ convolutions, which make them unsuitable for mobile applications that have strict power and latency constraints. Our network uses a backbone of only $2D$ convolutions with added Temporal Shift Modules (TSM) \cite{lin2019tsm} to facilitate an information exchange between temporally adjacent frames. This results in a light-weight network architecture that needs very limited memory and compute requirement. The proposed network architecture is shown in Fig.\ref{Fig:Architecture}. Our network is inspired by TSN \cite{wang2016temporal} architecture, where a set of frames is sampled from a video and processed independently. Finally, a consensus is taken to obtain a global feature representation. We first divide the input video temporally into $T$ segments of equal durations and one representative central frame is sampled from each segment. The input of our model is thus a $T * N * N$ volume, where $T$ is the number of segments from the video and $N$ is both the height and the width of the video. The input volume is passed through a TSN-style backbone network to obtain a $T * 1280$ shape feature representation. The obtained feature vector is then averaged over the temporal dimension to obtain a combined 1280-dimension feature vector for the entire video. This global video feature vector is then fed into a classifier head having two fully connected layers with $128$ and $64$ neurons respectively, followed by a softmax layer for classification.
The working of the original TSN architecture is explained by the equation~\ref{tsn_eq}. The video $V$ is divided into $K$ segments \{$S_1$,$S_2$, \dots, $S_K$\} of equal duration and ($T_1$,$T_2$, \dots, $T_K$) are the sequence of snippets where each $T_K$ is sampled from its corresponding segment $S_K$. $\mathcal{F}$($T_K$; $W$) defines the output after passing the snippet $T_K$ through the ConvNet with  parameters $W$. The consensus module $\mathcal{G}$ combines the extracted features of all the snippets through $\mathcal{F}$ operation. The consensus module for our architecture takes the average of the features. The average output of consensus module is passed through the fully-connected layer with a softmax at the end to get the final class label. This operation is defined by $\mathcal{H}$ in the equation~\ref{tsn_eq}.
\begin{equation}\label{tsn_eq}
\begin{split}
TSN(T_1, T_2, \dots, T_K) = \mathcal{H}(\mathcal{G}(\mathcal{F}(T_1;W),\\
\mathcal{F}(T_2;W),\dots, \mathcal{F}(T_K;W)))
\end{split}
\end{equation}

We added TSM modules in the backbone network to help the network learn strong temporal relations across segments via shifting of the intermediate feature channels of one segment to neighboring segments. To further reduce the computational complexity of our network, we have used MobileNetV2 \cite{sandler2018mobilenetv2} as the backbone due to its low computational cost. Our specific design choices for the network architecture makes it suitable for video processing on mobile devices having low compute budget.


\subsection{Video Playback Style Recommendation}
Applying a suitable playback style to a video can enhance a video and make it more likely to be shared.
Motion patterns present in the videos play an important role in selecting the most suited playback style for the video. For instance, for a video having linear motion like running, applying Boomerang type will make the video counter intuitive and hence interesting. To this end, we have designed a system for video playback style recommendation based on predictions from motion type classifier. We have considered three most widely used playback styles for recommendation namely Boomerang, Loop and Reverse. Specifically, we have introduced a mapping from motion type to a suitable playback style for an input video based on a user survey of 14 volunteers. In this study, we showed a few example actions for each motion type to each volunteer and asked them to select the best-suited playback style for that corresponding action. We aggregated the results from each volunteer and selected the most voted playback style for each motion type for the mapping. From the results of the study as shown in Table  \ref{tab:playback_mapping}, we observe that the Reverse effect suits linear actions, and projectile motion looks good with a Boomerang effect. For both oscillatory and local motion, loop is the best-suited playback style. For random motion type, we randomly apply  Boomerang, Reverse or Loop. We have performed a subjective study for evaluation of our video playback style recommendation system which is detailed in Sec. \ref{subsec:playback_exp}.

\section{Experiments}

We have done multiple experiments for comprehensive evaluation of our proposed motion type classifier model. In Section \ref{subsubsec:motion_classifier} we perform an ablation with various pre-trained weights to examine the impact of weight initialization. To evaluate the quality of the learnt representations through motion classification, we have performed video retrieval as detailed in Sec. \ref{subsec:vid_retrieval}. We have also performed a subjective study for evaluation of our video playback recommendation system. To prepare our training data, each video was first resized: the smaller dimension was set to $256$ pixels wide, and a random square region was cropped of side length $d$ $\epsilon$ $(256, 224, 192, 169)$ , followed by a random horizontal flip. Finally, the crop was resized to $(256, 256)$ and the pixel values were normalized to the range $(0, 1)$. In the testing phase, we resized the smaller dimension to $256$ and took a center crop. We used $T=3$ segments in all of our experiments unless mentioned otherwise, and sampled the temporally central frame from each segment. These three frames are the input to the network. For training, we used an initial learning rate of $0.001$ and a learning rate schedule to reduce the learning rate by half after the $20^{th}$ and $40^{th}$ epoch. The network was trained for a total of $200$ epochs. Stochastic Gradient Descent was used for optimization with momentum value of $0.9$ and a weight decay of $5e-5$. We have trained all our models with a single P100 GPU and each training configuration took 4hrs to converge.

\subsection{Dataset}\label{subsec:dataset}
For all our experiments, we use the HMDB$51$ \cite{kuehne2011hmdb} dataset. The HMDB$51$ dataset contains short videos (1-15 seconds) for $51$ human actions like cycling, eating, running and dancing etc. We have used the split-1 set of HMDB$51$ provided by \cite{kuehne2011hmdb} to create the train/test/validation set. There are $3570$ videos in the train set, $1530$ videos in the test set and $1749$ in the validation set. These videos are collected from YouTube and digitized movies and have large variability in camera motion, view-point and illumination. For our purpose, we annotated each of the $51$ action classes from the HMDB$51$ dataset with one of our five defined motion types. We have named this annotated version of the HMDB$51$ dataset the mHMDB$51$ dataset. A subset of this mapping is shown in Table~\ref{tab:playback_mapping}, while the full version can be found in the appendix.


\subsection{Motion Classifier} \label{subsubsec:motion_classifier}
For evaluation, we have compared our model with a optical flow based baseline model and performed an ablation study with various pre-training methods. The results are shown in Table \ref{tab:motion_acc}.

\subsubsection{Baseline Classifier}\label{subsubsec:flow} To benchmark our motion type classifier, we designed a baseline classifier as a two-layer fully connected neural network. The input to this classifier is based on the statistics of motion magnitudes in the video. To extract the input features, we first compute the pixel-wise average over time of the motion boundaries for the input video and divide it into $16$ cells as in \cite{wang2019self}. We use the standard deviation of the magnitude of motion boundaries within each cell to form the 16-dimensional input feature vector to the motion type classifier. In the network design, there are $128$ neurons in the first hidden layer and $5$ neurons in the second hidden layer for the network. ReLU activation was used after the first hidden layer and a softmax activation was applied after the second hidden layer for the final classification. Dropout regularization was applied with a drop probability of $0.2$ and the classifier was trained for $5$ epochs with a learning rate of $0.001$.

\subsubsection{Model Performance Analysis}
\begin{figure*}[t]
  \centering
  \includegraphics[width=16.5cm, height=11.5cm]{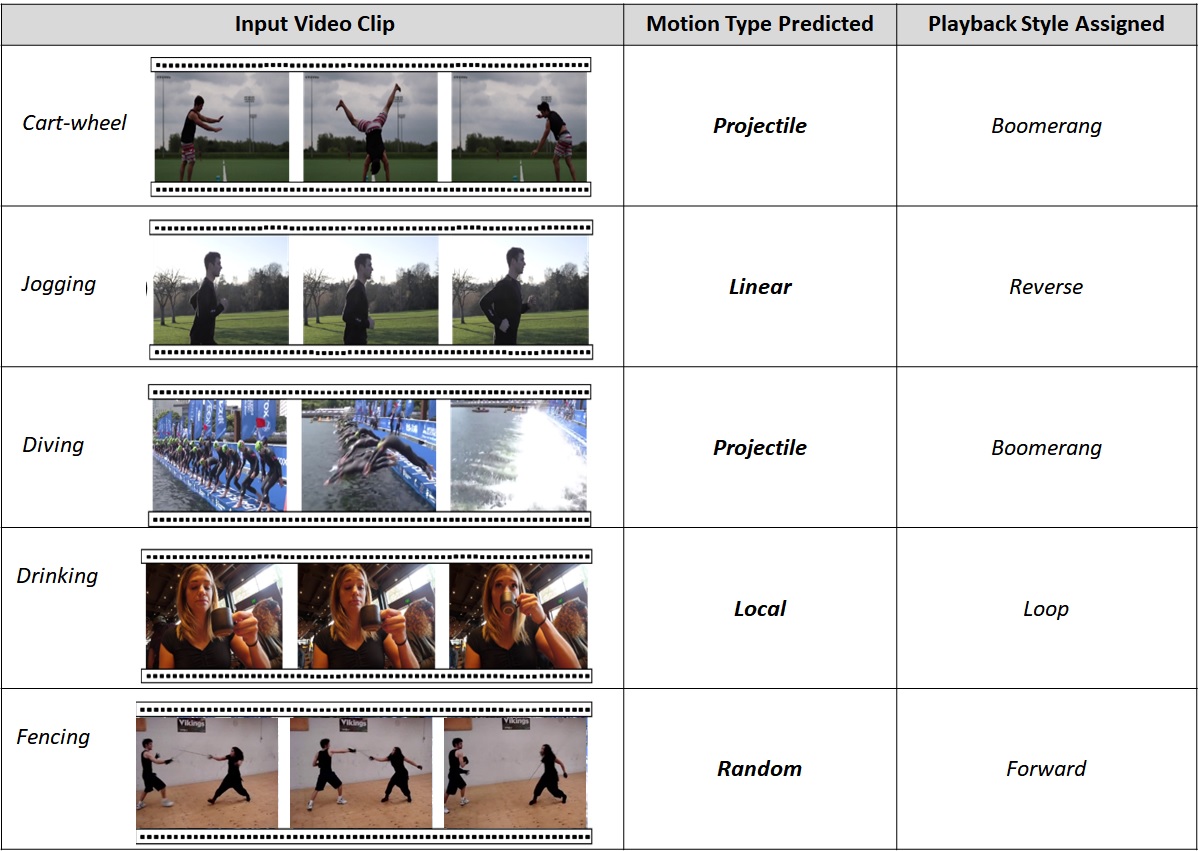}
    \caption{Playback style recommendation by our system for YouTube videos}
    \label{Fig:playback_youtube}
\end{figure*}
We observed that training our classifier from scratch achieved a performance boost of nearly $13\%$ over the baseline flow-based model, but still low as compared to fully supervised pre-training with ImageNet \cite{deng2009imagenet} and Kinetics \cite{kay2017kinetics}. This was expected behavior, as our model was trained with only $3500$ videos from the HMDB$51$ dataset, which is insufficient for supervised training when compared to the millions of data points used to train existing ImageNet and Kinetics classifiers. Thus our usage of transfer learning via initializing our classifier with weights learned from the ImageNet classification task increased our accuracy by a margin of around $14\%$, due to the pre-trained understanding of important spatial features. Initializing with weights learned for action classification on the Kinetics dataset achieved the best accuracy, as they have a pre-trained understanding of both spatial and temporal features, which are useful to perform motion classification. Our baseline local-flow-based classifier expectedly performed the worst. These observations indicate that accurately predicting object motion type requires global semantic information contained in motion patterns. our results demonstrate that our motion type classifier learns more than just the motion magnitude, and has a deeper understanding of object motion patterns.

\subsubsection{Model Complexity Analysis}
We also performed an ablation study by varying the complexity of the backbone network. Our baseline model is TSN \cite{wang2016temporal} with shift modules which process multiple segments from a video and fuse them together at a later stage to obtain the combined feature vector. As the number of segments represent the complexity of the model, we have trained models with 1, 2, 3 and 8 segments in our ablation study. The overall accuracy of the model and the number of multiply-accumulate (MAC) operations is shown in Table \ref{tab:segment_comparison}. The three-segment model achieved the best accuracy for motion type classification. However, the two-segment model was able to achieve comparable accuracies to the three-segment model with just 0.82G MAC operations, making it the optimally suited configuration for mobile deployment. The inference time for the two-segment model on a Samsung S20 mobile device running a Qualcomm Snapdragon Adreno 650 GPU is just 200 milliseconds.
The single-segment model processes only a single frame from the complete video and therefore struggles to learn temporal dynamics of the video. However, it was still able to achieve a reasonable accuracy of $61.75\%$ for motion type classification, demonstrating the importance of object appearance in determining the natural motion patterns for an object. The eight-segment model did not perform well, due to the HMDB$51$ dataset having small action videos and thus not requiring too many frames for effective motion pattern understanding. We believe that passing a large number of frames for actions with short duration captures multiple motion types present in the video at different instances and hence confuses the network training.

\begin{table}[t]
\begin{center}
\caption{Motion Type Classifier Top-1 Accuracy.} \label{tab:motion_acc}
\begin{tabular}{|l|l|}
  \hline
  Method & Accuracy  \\
  \hline
  Baseline Classifier & 25.64 \\
  $\text{Ours}_{\text{Scratch}}$ & 38.56 \\
  $\text{Ours}_{\text{ImageNet}}$ & 57.58  \\
  $\text{Ours}_{\text{Kinetics}}$ & 72.68 \\
  \hline
\end{tabular}

\vspace{-0.1cm}
\end{center}
\end{table}
\begin{figure}[b]
  \centering
  \includegraphics[width=8cm, height=3cm]{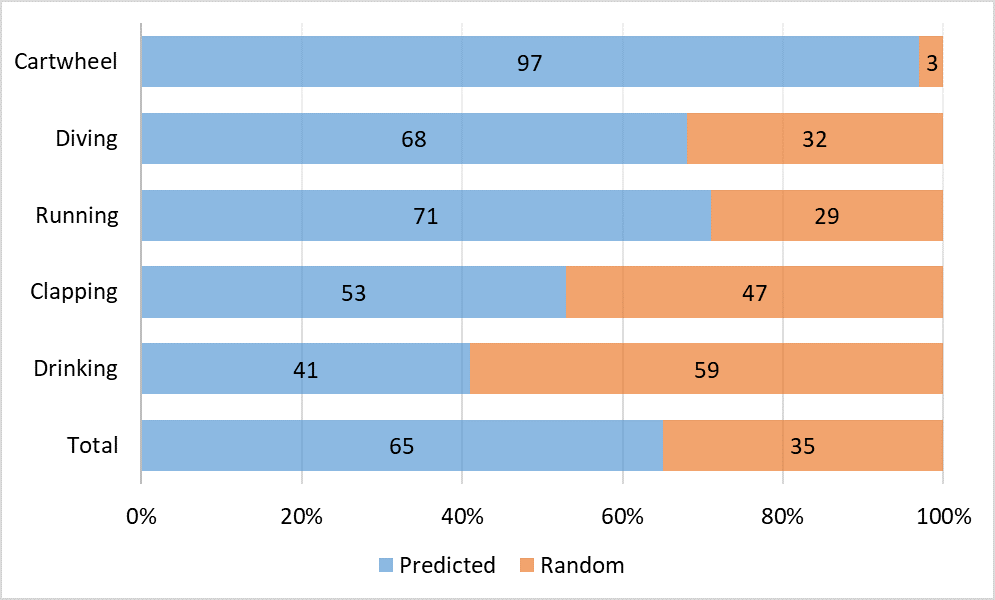}
    \caption{Subjective study for video playback style recommendation.}
    \vspace{-0.1cm}
    \label{Fig:subjective_study}
\end{figure}

\subsection{Video Playback Style Recommendation}\label{subsec:playback_exp}
\begin{table}[t]
\begin{center}
\caption{Comparison of motion classifier top-1 accuracy and MAC operations for a varying number of input segments for the network.} \label{tab:segment_comparison}
\begin{tabular}{|l|l|l|}
    \hline

  Segments & Accuracy & MACs \\
  \hline
  1 & 61.76 & 0.41G \\
  2 & 71.05 & 0.82G \\
  3 & 72.68 & 1.23G \\
  8 & 68.17 & 3.28G \\

  \hline
\end{tabular}
\vspace{-0.2cm}
\end{center}
\end{table}
For a subjective evaluation of our video playback style recommendations, we conducted a user study with $10$ volunteers. We downloaded two clips for each of the following five actions from YouTube: \emph{cartwheel, diving, running, clapping, and drinking}. Our network predicted the motion type of each video, and we applied the matching playback style based on the mapping shown in Table \ref{tab:playback_mapping}. We also prepared a comparison set for the same videos with randomly applied playback styles. We evaluated our recommended playback styles against these randomly selected playback styles. The volunteers were asked to select the most aesthetic and preferred result from these two sets, the results of which are shown in Fig. \ref{Fig:subjective_study}. For the categories that have a large global motions like cartwheel, diving, and running, our predicted playback style was ranked better than random playback style on an average. To our surprise, while the diving action was not present in our training set, our engine was able to recommend the best-suited playback style for the class. This provides evidence for the proposition that training to predict motion type captures more abstract information than actions, and generalizes well for unseen data. On the contrary, for local action categories such as drinking and clapping, our method was indistinguishable to random selection as the impact of playback style is not very evident when motion is confined to a small spatial region.

\subsection{Video Retrieval}\label{subsec:vid_retrieval}
\begin{figure*}[t]
  \centering
  \includegraphics[width=16cm, height=14cm]{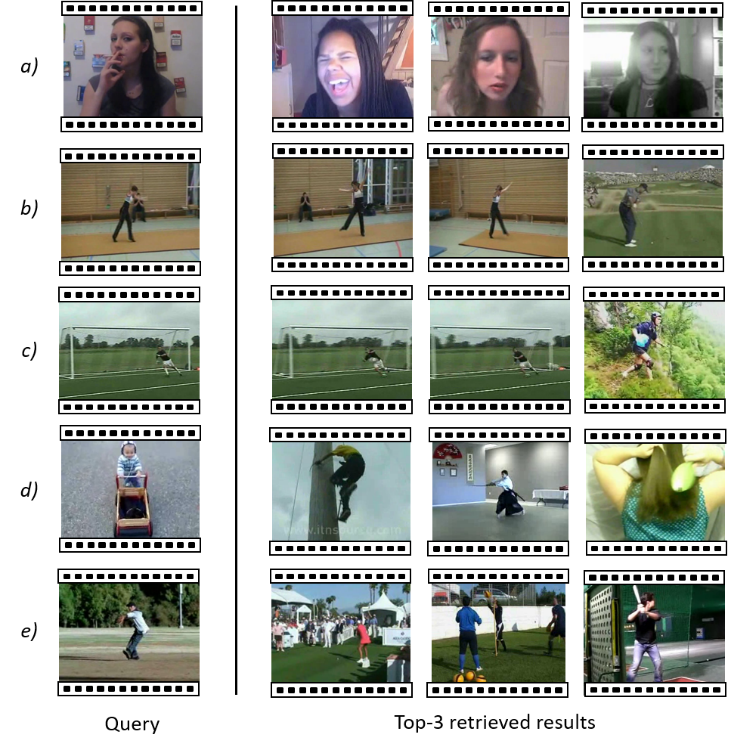}
    \caption{Video retrieval from HMDB51 dataset using learned feature vectors}
    \label{Fig:Video retrieval}
\end{figure*}

To further analyze the spatio-temporal features learned by our motion type classifier, we used these features to perform video retrieval. Given a query video, we aim to find the three most similar videos to the query video from a database of videos. We feed all the videos from HMDB$51$ to our motion classifier and extract the $1280$-dimensional feature vector described in Sec.~\ref{sec:arch} for each video. In an ideal scenario, this feature vector represents the motion present in the video in a compressed form. We apply the k-nearest-neighbor algorithm in the $1280$-dimensional feature vector space to find videos having similar motion patterns as that of the query video. Some example retrievals from HMDB$51$ are shown in Fig.~\ref{Fig:Video retrieval}, from which it is evident that our learned representations capture meaningful semantic information of object motion. In Fig.~\ref{Fig:Video retrieval}a) the query video was of smoking, and all retrieved results (laugh, chew and chew) have local facial motions. In Fig.~\ref{Fig:Video retrieval}b) and c) the first two results are from the same scene but at different points in time. In Fig.~\ref{Fig:Video retrieval}b) the third retrieved result is of a golf swing, which has similar hand movement to that of a cartwheel. Similarly, for c) the last retrieved result is of a person diving from a cliff, which is very similar to the query video of a goalkeeper diving for football. For Fig.~\ref{Fig:Video retrieval}d) all retrieved videos have linear motion and in Fig.~\ref{Fig:Video retrieval}e) all the retrieved actions for the query video of throw follow projectile motion.

\section{Conclusion}
In this work, we have examined the importance of object motion features in video analysis. We trained a model that understands the underlying object motion patterns and classifies the object motion into one of the five defined directional motion classes. We have also shown the exciting use case of playback style recommendation based on our classifier's predicted motion type. Finally, we have evaluated the representations learned by motion type classifiers for video retrieval and have found that these representations generalize well for this task. In the future, we plan to explore other possible approaches to model object motions in the videos. We will also evaluate the generalization ability of learned representations for more challenging video tasks such as action localization and classification.

{\small
\bibliographystyle{ieee_fullname}
\bibliography{egpaper_final_main}
}

\appendix
\newpage
\section{Appendix}
\subsection{Action to Motion Type Mapping}
We have manually annotated all the action classes present in HMDB51 with motion type classes based on the mapping shown in Table \ref{tab:action_motion_mapping}. to obtain mHMDB51 dataset. We have used mHMDB51 dataset for training and evaluation of motion type classifier.

\begin{table}[t]
\begin{center}
\caption{Motion Type mapping based on action class} \label{tab:action_motion_mapping}
\begin{tabular}{|l|l|}
  \hline
  Action & Motion Type \\
  \hline
    brush\_hair & Linear \\
    cartwheel & Projectile \\
    catch & Projectile \\
    chew & Local \\
    clap & Oscillatory \\
    climb & Linear \\
    climb\_stairs & Linear \\
    dive & Projectile \\
    draw\_sword & Random \\
    dribble & Oscillatory \\
    drink & Local \\
    eat & Local \\
    fall\_floor & Random \\
    fencing & Random \\
    flic\_flac & Projectile \\
    golf & Projectile \\
    handstand & Projectile \\
    hit & Projectile \\
    hug & Random \\
    jump & Projectile \\
    kick & Random \\
    kick\_ball & Random \\
    kiss & Local \\
    laugh & Local \\
    pick & Random \\
    pour & Local \\
    pullup & Oscillatory \\
    punch & Linear \\
    push & Linear \\
    pushup & Oscillatory \\
    ride\_bike & Linear \\
    ride\_horse & Linear \\
    run & Linear \\
    shake\_hands & Local \\
    shoot\_ball & Projectile \\
    shoot\_bow & Linear \\
    shoot\_gun & Local \\
    sit & Random \\
    situp & Oscillatory \\
    smile & Local \\
    smoke & Local \\
    somersault & Projectile \\
    stand & Random \\
    swing\_baseball & Projectile \\
    sword & Random \\
    sword\_exercise & Random \\
    talk & Local \\
    throw & Projectile \\
    turn & Random \\
    walk & Linear \\
    wave & Local \\
    \hline
\end{tabular}
\end{center}
\end{table}

\end{document}